# XEngine: Optimal Tensor Rematerialization for Neural Networks in Heterogeneous Environments

MANUELA SCHULER, Deutsches Forschungszentrum für Künstliche Intelligenz (DFKI), Saarland Informatics Campus, Germany

RICHARD MEMBARTH, Technische Hochschule Ingolstadt, Research Institute AImotion Bavaria, Germany and Deutsches Forschungszentrum für Künstliche Intelligenz (DFKI), Saarland Informatics Campus, Germany

PHILIPP SLUSALLEK, Deutsches Forschungszentrum für Künstliche Intelligenz (DFKI), Saarland Informatics Campus, Germany and Saarland University, Saarland Informatics Campus, Germany

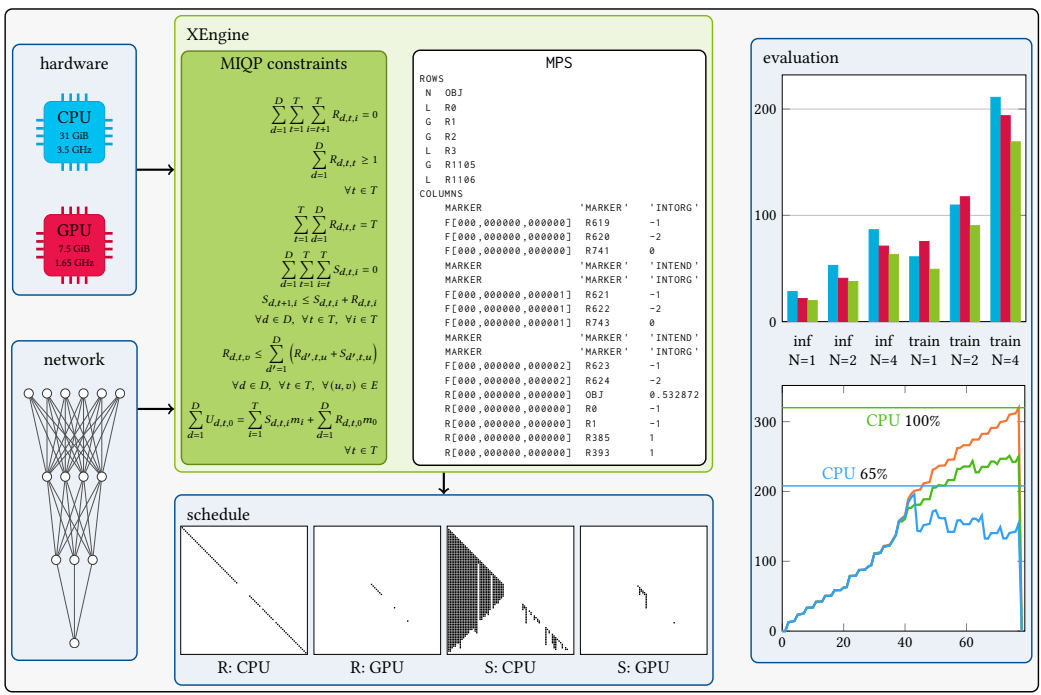

Fig. 1. XEngine reads the hardware and network topology, defines constraints and generates the MIQP as MPS file. It then solves the MIQP and outputs a schedule for all devices (second figure in the middle). The evaluation involves measuring the compute performance (first figure on the right side) as well as the used memory of the schedule compared to an save-all-tensors-strategy.

This work is supported by the Federal Ministry of Education and Research (BMBF) as part of the ENGAGE, HorME, HP-DLF, and MetaDL projects.

Authors' addresses: Manuela Schuler, Deutsches Forschungszentrum für Künstliche Intelligenz (DFKI), Saarland Informatics Campus, Saarbrücken, Germany, 66123, manuela.schuler@dfki.de; Richard Membarth, Technische Hochschule Ingolstadt, Research Institute AImotion Bavaria, Ingolstadt, Germany, 85049 and Deutsches Forschungszentrum für Künstliche Intelligenz (DFKI), Saarland Informatics Campus, Saarbrücken, Germany, 66123, richard.membarth@thi.de; Philipp Slusallek, Deutsches Forschungszentrum für Künstliche Intelligenz (DFKI), Saarland Informatics Campus, Saarbrücken, Germany, 66123 and Saarland University, Saarland Informatics Campus, Saarbrücken, Germany, 66123, philipp.slusallek@dfki.de.

Memory efficiency is crucial in training deep learning networks on resource-restricted devices. During backpropagation, forward tensors are used to calculate gradients. Despite the option of keeping those dependencies in memory until they are reused in backpropagation, some forward tensors can be discarded and recomputed later from saved tensors, so-called checkpoints. This allows, in particular, for resource-constrained heterogeneous environments to make use of all available compute devices. Unfortunately, the definition of these checkpoints is a non-trivial problem and poses a challenge to the programmer—improper or excessive recomputations negate the benefit of checkpointing.

In this article, we present XEngine, an approach that schedules network operators to heterogeneous devices in low memory environments by determining checkpoints and recomputations of tensors. Our approach selects suitable resources per timestep and operator and optimizes the end-to-end time for neural networks taking the memory limitation of each device into account. For this, we formulate a mixed-integer quadratic program (MIQP) to schedule operators of deep learning networks on heterogeneous systems. We compare our MIQP solver XEngine against Checkmate [12], a mixed-integer linear programming (MILP) approach that solves recomputation on a single device. Our solver finds solutions that are up to 22.5% faster than the fastest Checkmate schedule in which the network is computed exclusively on a single device. We also find valid schedules for networks making use of both central processing units and graphics processing units if memory limitations do not allow scheduling exclusively to the graphics processing unit.

CCS Concepts: • **Computer systems organization** → **Neural networks**; Heterogeneous (hybrid) systems; • **Computing methodologies** → Distributed algorithms; • **Theory of computation** → Integer programming; Scheduling algorithms.

Additional Key Words and Phrases: rematerialization, integer linear programming, neural networks, memory management, heterogeneous computing

ACM Reference Format:
Manuela Schuler, Richard Membarth, and Philipp Slusallek. 2022. XEngine: Optimal Tensor Rematerialization for Neural Networks in Heterogeneous Environments. *ACM Trans. Arch. Code Optim.* 20, 1, Article 17 (December 2022), 26 pages. https://doi.org/10.1145/3568956

## 1 INTRODUCTION

Memory is one of the most limiting factors in training deep neural networks [3, 4, 12]. In a training step, forward activations are usually cached and reused during the backward pass. In case of insufficient memory to cache all forward activations, some tensors can be discarded. When needed for gradient calculation, they can be recomputed from stored tensors called checkpoints. This way, models can fit into memory at the expense of additional compute costs. Since compute performance is also often limited in low-memory environments, it is crucial to make best use of resources. In this work, we consider all available devices and combine the idea of tensor rematerialization with distributed computation and tensor swapping in order to get the best performance. Manually assigning the operators to different devices, even without recomputation, is hard—dependency structures can get quite complicated, especially in the backward pass. Simply scheduling the operator on the device with highest compute performance often yields suboptimal schedules due to high copy costs. Our approach tackles the problem of scheduling network operators to heterogeneous compute devices with different compute capabilities and different memory limits using a mixed-integer quadratic programming (MIQP) approach.

There are various approaches in literature that address the rematerialization of tensors. Earlier works in the field focus on rematerialization of graph segments [3, 5–8, 24]. Graph-theoretic analyses achieve better results by rematerialization of individual components [1, 12, 14–16]. All of these approaches consider only a single-device setup. As rematerialization enables training on devices with limited memory, it can also help to address the problem of scheduling the network operators to various heterogeneous devices.

To the best of our knowledge, there is no other work at the moment that combines rematerialization with distributed computation on multiple heterogeneous devices.

We formalize the scheduling problem as an MIQP, using Checkmate [12] as a starting point for our work. This article makes the following contributions:

- We solve the resource selection and rematerialization problem for inference and training of neural networks on heterogeneous devices using an MIQP.
- We show that if a central processing unit (CPU) and graphics processing unit (GPU) have a similar compute performance in a heterogeneous setting, CPU/GPU schedules can outperform single-device schedules.
- We achieve valid CPU/GPU schedules where computation on only the GPU device would exceed memory limitations.
- We show how the MIQP can be extended to also consider energy efficiency.

Our experiments show that if the CPU's performance is comparable to that of the GPU, our approach is clearly faster than computing everything only on the GPU.

We gain all relevant information from running the network operators on all devices in a first step, followed by acquiring copy costs between the devices for all tensors. Using this simple cost model, we formulate the MIQP and solve it. The solution is translated into a detailed plan of computing, saving, or freeing the output tensors of operators.

XEngine is open source and available on GitHub: https://github.com/dfki-asr/xengine.

Since all operator and network copy costs are measured beforehand in our pipeline, it is not possible to adapt to online changes (adding or removing resources during training). All information has to be known at compile time. We support static graphs only and do not consider dynamic graph architectures.

## 2 RELATED WORK
### 2.1 Rematerialization and Checkpointing

Checkpointing in deep learning has its roots in Griewank's Revolve approach [5, 6] for reversemode Automatic Differentiation, where some activations in the "tape" are used to recompute other previously discarded values. Hascoët and Pascual [8] and Siskind and Pearlmutter [24] extend this idea to handle arbitrary control flow with policies known at compile time. Chen et al. [3] treat neural networks as static graphs and divide them into segments to be recomputed during backpropagation. Gruslys et al. [7] expand this segmenting approach to recurrent neural networks.

Graph-theoretic analyses [1, 15, 16] or mixed-integer linear programming (MILP) such as Checkmate [12] achieve better bounds with rematerialization of individual activations rather than entire segments.

Kirisame et al. [14] use heuristics in the form of a greedy online algorithm. Rematerialization is triggered as soon as memory is exhausted and parent operators are recursively recomputed.

Beaumont et al. [2] combine rematerialization with memory offloading using dynamic programming.

All of these approaches focus on computation on a single device rather than scheduling the computations on multiple devices.

Hu et al. [10] present MegTaiChi, a system that is tracking fine-grained tensor accesses and focuses on dynamic tensor partitioning and memory fragmentation and defragmentation. Compared with our approach, they use an online approach to also handle dynamic graphs, which introduces additional overhead per iteration. They claim that their runtime overhead is less than 5% of the runtime of each iteration. Our approach does not introduce additional online overhead since the schedule is computed beforehand. For static graphs, an offline approach will be sufficient and an online approach will only introduce additional overhead. For dynamic graph applications, an online approach is strictly necessary since the graph changes during iterations dependent on the input data and the runtimes cannot be acquired beforehand in an offline step. A recent work of Liao et al. [17] proposes Mimose, an input-aware dynamic planner composed of an online collector, a regression-based memory estimator, and a memory scheduler that mainly focuses on GPU devices. The architecture of DELTA, a system proposed by Tang et al. [25], consists of a heuristic-based filter component, a director, and a prefetcher component, which also focuses on dynamic graphs.

## 2.2 Memory Offloading, Swapping

Swapping allows training of neural networks in low-memory environments by offloading memory from the accelerator to the host device. Huang et al. [11] use a generic algorithm to train models up to 12 times the GPU limit by using smart swapping and still get 53% to 99% of the throughput compared with infinite GPU memory. Capuchin [20] and Superneurons [27] combine rematerialization and swapping by gathering network information in a runtime system. Capuchin gathers swapping cost information during a single batch run to determine where to set checkpoints. Beaumont et al. [2] use dynamic programming to solve rematerialization with memory offloading. Patil et al. [19] present POET (Private Optimal Energy Training), a MILP-based system that exploits the off-chip memory for energy-efficient paging on battery-operated edge devices. Their work combines memory offloading with an MILP-based rematerialization approach. ZeRO-Offload, a work of Ren et al. [21], is a CPU-GPU framework that uses a trick of "One-step Delayed Parameter Update (DPU)" to achieve scaling of the GPU throughput. Wen et al. [28] propose a swap dominated tensor re-generation strategy, called STR, to avoid negative effects of improper swap decisions when the source of the recomputation may have been swapped out.

These approaches assume that computation exclusively takes place on the GPU and offload the memory only to the CPU.

## 2.3 Distributed Computation and Computation Offloading

Offloading computations to other devices can overcome the limited memory and compute performance issue of single devices such as mobile phones. To make offloading decisions, Liu et al. [18] formulate a linear program while Van Le and Tham [26] use reinforcement learning. Jiang et al. [13] combine the compute performance of smartphone devices with a cloud server to distribute the training of neural networks.

We are not going to offload computations to remote devices; rather, we will distribute computations within only one node between the CPU and GPU.

## 3 REMATERIALIZATION WITH MULTIPLE DEVICES
## 3.1 Problem Definition

The rematerialization approach of Checkmate [12] considers computation only on the GPU device. When the CPU and GPU have a comparable performance, it can be beneficial to consider the CPU as a second device for computation, not only for offloading tensors.

In this work, we present XEngine, an MIQP-based rematerialization approach for multiple devices in heterogeneous systems such as CPUs and GPUs. We extended the MILP of Checkmate towards multiple heterogeneous devices by adding additional constraints and dimensions. We evaluate

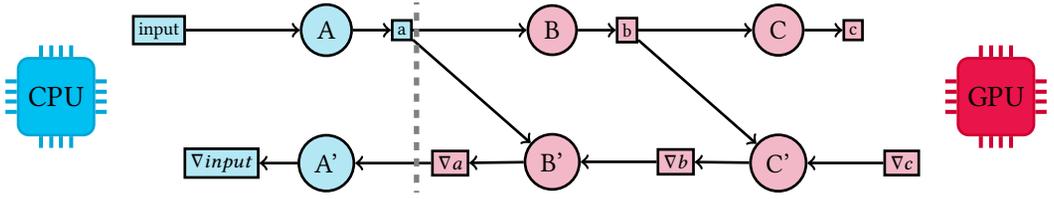

Fig. 2. Forward operators $A$, $B$, $C$ with output tensors $a$, $b$, $c$. Backward operators $A'$, $B'$, $C'$ with output tensors $\nabla input$, $\nabla a$, $\nabla b$. $A$ and $A'$ are assigned to the CPU (blue), all other operators to the GPU (red). Tensor $a$ (blue) must be copied from the CPU to the GPU twice: once in the forward path to compute $B$ and once in the backward path to compute $B'$. $\nabla a$ must be be copied from the GPU to the CPU to compute $A'$ (blue).

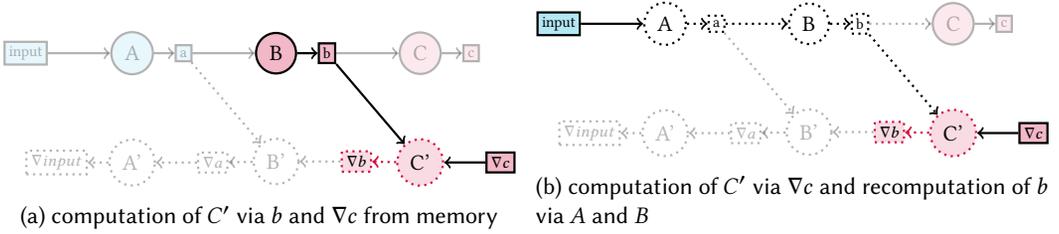

(a) computation of $C'$ via $b$ and $\nabla c$ from memory

(b) computation of $C'$ via $\nabla c$ and recomputation of $b$ via $A$ and $B$

Fig. 3. (a) Direct computation of operator $C'$ (red dotted circle), since tensors $b$ and $\nabla c$ are in memory (solid rectangles). (b) Computation of operator $C'$ (red dotted circle) triggers recomputation of operators $A$ and $B$ (dotted circles) since $C'$ directly depends on tensor $b$ and indirectly on tensor $a$, both not located in memory (dotted rectangles).

our solver considering distribution and recomputation on multiple devices against the Checkmate solver, which considers recomputation on only a single device.

We combine two aspects in our work: the distribution on devices and rematerialization of tensors.

*3.1.1 Distribution on Devices.* We consider multiple devices for computation of the network operators. Figure 2 shows the distribution aspect: a simple example network with three forward and three backward operators (circles) is distributed on a CPU and a GPU. Operators that are scheduled to the CPU are marked blue and those computed on the GPU are marked red. Their corresponding output tensors are denoted as blue (on the CPU) or red (on the GPU) rectangles. The first forward operator $A$ is scheduled on the CPU device, whereas the network forward operators $B$ and $C$ are scheduled on the GPU. Their corresponding backward operators $B'$ and $C'$ are scheduled on the GPU, whereas $A'$ again is put on the CPU device. Device switches at the boundaries cause tensor copies between the devices: tensor $a$ and $\nabla a$ have to be copied for this transition.

*3.1.2 Rematerialization of Tensors.* In low-memory environments, tensors can be discarded in the forward pass and recomputed from checkpoints in the backward pass. Figure 3 shows the recomputation mechanism: any dependent tensor that is not available in memory must be rematerialized. In case the tensor is located on another device, the tensor can be either copied or rematerialized via computation on the same device. Operators and tensors located in memory are marked solid, whereas dotted marking denotes that the operator or tensor is not located in memory. In the example, the point in time is depicted where operator $C'$ (red dotted markings) is computed on the GPU to yield output tensor $\nabla b$ (red dotted markings).

Figure 3(a) shows the case without rematerialization: all direct dependencies (tensors $\nabla c$ and $b$) of operator $C'$ are available in memory (solid circles). Thus, no recomputation is necessary. Since tensor $b$ is already located on the GPU, it does not have to be copied from the CPU; it can be used directly for computation of $C'$.

Figure 3(a) shows a situation in which tensor $b$ is not available in memory (dotted circle). It might have been freed after computation of the forward operator $C$. Since $C'$ directly depends on $b$, operator $B$ has to be recomputed before we can compute $C'$. $B$ itself depends on tensor $a$, which is also not available and must be rematerialized as well. Tensor $b$ can be rematerialized starting from the *input* tensor (blue), which is located in CPU memory. Before $C'$ can be computed, operators $A$ and $B$ have to be recomputed on either the CPU or GPU, accounting for compute costs and copy costs for potential device switches.

There are three different options:

- Copy *input* to GPU, compute $A$ on GPU, compute $B$ on GPU, compute $C'$ on GPU
- Compute $A$ on CPU, copy $a$ to GPU, compute $B$ on GPU, compute $C'$ on GPU
- Compute $A$ on CPU, compute $B$ on CPU, copy $b$ to GPU, compute $C'$ on GPU

In order to visualize our schedules, we use a technique similar to that of Checkmate [12]. We introduce two matrices $R$ and $S$, each of shape $|T \times T|$: The number of rows in each matrix corresponds to the number of timesteps $T$. Since in every timestep exactly one new operator is computed, we need $T$ timesteps to compute every operator at least once. Therefore, $T$ is equal to the number of operators in the network graph.

The recompute matrix $R$ shows operator computations (black square denotes "compute"). The save matrix $S$ shows whether the output of an operator is saved after computation (black square denotes "save").

We define the memory budget needed to save all tensors as 100%. For training a UNet with batchsize 2, 320 MiB of memory is needed when no tensor is discarded and every tensor is saved. UNet architectures were introduced by Ronneberger et al. [22] and contain skip connections that introduce tensor dependencies in combination with high memory demand. Figure 4 shows our XEngine schedule for training this UNet on the CPU and GPU with a restricted memory limit of 208 MiB (65%) on each device. We show the matrices $R$ and $S$ for both devices: $R_0$ (compute CPU), $R_1$ (compute GPU) and $S_0$ (save CPU), $S_1$ (save GPU).

Columns correspond to the operator index and rows to the timestep. The matrices are read from the left to right and top to bottom: most of the operators are scheduled to the CPU ($R_0$, first figure on left, Figure 4). Some operators, especially at the end of the forward and start of the backward pass, are computed on the GPU ($R_1$, second figure from left, Figure 4). Each operator is computed only once, either on the CPU or on the GPU. The left lower part of the diagonal does not show any recomputation in both matrices $R_0$ and $R_1$. The save matrix of the CPU $S_0$ (third figure from left, Figure 4) shows on the left half of the matrix that many forward activations are kept in memory until they are finally reused in the backward pass.

3.1.3 *Memory Usage of Schedules.* Figure 5 shows the memory usage for training a UNet with batchsize 2 for 1 iteration. We measure the total memory usage on each device after each timestep.

Three different cases are depicted:

- We visualize the memory usage for the case in which no tensor is freed during the iteration as an orange curve in both figures (keep all tensors). This corresponds to a schedule of a deep learning framework that does not release any intermediate memory during training iterations.
- We show the Checkmate schedule (CPU) for 100% budget and 65% budget. The green curve in Figure 5(a) shows the memory usage for the Checkmate schedule when the UNet is trained on the CPU with a sufficient memory limit of 320 MiB (100%, green line). The green curve is

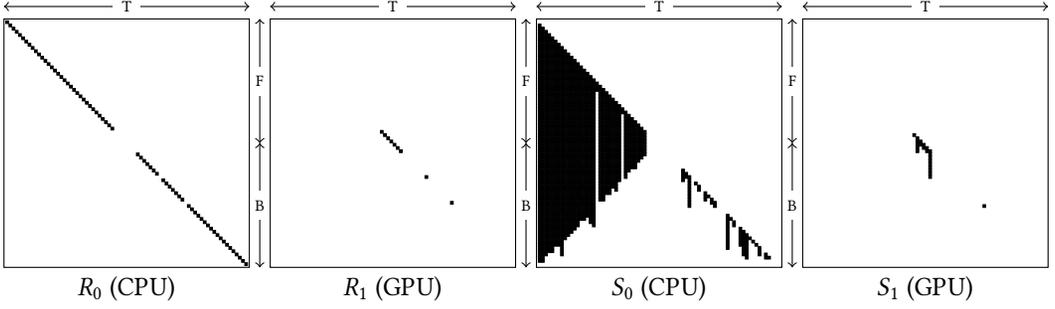

Fig. 4. XEngine schedule for training UNet with batchsize N=2 with 24 CPU threads and a memory budget of 208 MiB (65%) per device for the CPU and GPU. From left to right: compute matrix on the CPU $R_0$, compute matrix on the GPU $R_1$, and save matrices for the CPU $S_0$ and GPU $S_1$. The memory usage of this schedule is depicted in Figure 5b. F=forward pass, B=backward pass, T=operators.

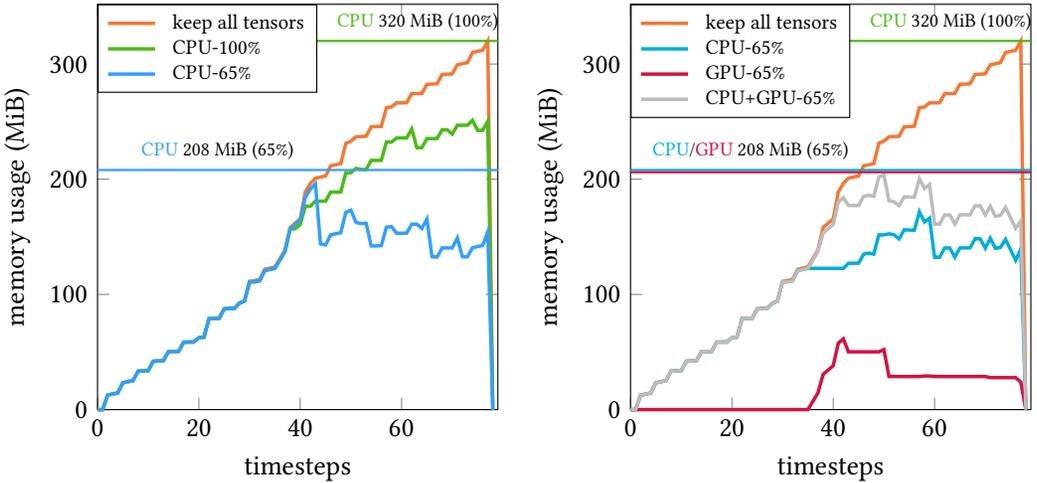

(a) Memory demand of Checkmate schedule for training UNet with N=2 on the CPU with a budget of 100% (green) and 65% (blue).

(b) Memory demand of XEngine schedule for training UNet with N=2 on the CPU and GPU with a budget of 208 MiB (65%) each.

Fig. 5. Memory usage for training UNet with batchsize N=2. The orange curve (keep all tensors) shows the memory usage per timestep if no tensor is freed over time during the whole iteration. (a) Checkmate schedule on the CPU for 320 MiB (100%, green) and 208 MiB (65%, blue) memory budget. (b) XEngine schedule on CPU (blue) and GPU (red) for 208 MiB (65%) memory budget for each device. The gray curve (CPU+GPU-65%) shows the total memory (on CPU and GPU) per timestep. (see XEngine schedule in Figure 4).

lower than the orange curve: some tensors in the backward pass can be freed directly after computation since they are no longer needed by any consecutive operator in the network. No tensor has to be recomputed. Lowering the budget to 208 MiB (65%, blue line) reduces the required memory further, which can be seen in the blue curve in Figure 5(a).

- We show the XEngine schedule (CPU, GPU) for 65% budget: Figure 5(b) shows the memory usage for the XEngine schedule in Figure 4 at a memory budget of 208 MiB (65%) per device. The blue curve shows the CPU memory over time, whereas the GPU memory is visualized as a red line. Both curves meet the required memory limit of 208 MiB each. The gray curve shows the total memory used in the system per timestep.

## 3.2 MIQP Formulation

Our aim is to schedule $T$ distinct operators in $T$ timesteps over $D$ devices such that minimum costs are achieved. We evaluate exactly one new operator per timestep $t$ and optionally recompute others. The binary matrices $R$, $S$, and $F$ denote (re-)computations, saves, and frees of operator outputs. $U$ is a continuous matrix that keeps track of the occupied memory. We introduce a new copy cost term $W$ to model device switches, considering the costs of tensor copies between devices. Each device $d \in D$ has its own computational costs per network operator $c_{d,i}$ and its own memory budget $b_d$. The solution of the MIQP allows recomputations and copies of tensors between devices as long as no memory budget of any device $b_d$ is exceeded. For simplicity, we consider only two distinct devices in most of our experiments: a CPU device as $d_0$ and a GPU device as $d_1$. However, we also show that the MIQP formulation can support more than two devices—for a CPU with 2 GPUs, for example.

*3.2.1 The Objective.* Our MIQP objective (Equation 1) contains the linear compute costs $c_{d,i}$ of operator $i$ on device $d$ and the new quadratic copy cost term $W$, which adds costs for copying tensors when devices are switched. $E$ is the set of all edges connecting the operators: $e \in E$ with $e = (u \rightarrow v)$ are tensors. $w_{e,d,d'}$ denotes the costs for copying tensor $e = (u \rightarrow v)$ from device $d$ to $d'$. These costs might be distinct from those copying the tensor back from $d'$ to $d$. Copy costs of tensors between devices are profiled together with computational costs on the device in an offline step and used in the MIQP formulation. We introduce matrix $Z$ to define where device switches can occur. We also extend every constraint of the Checkmate MILP by the new device dimension $d$.

$$\arg\min_{R,S,U,F,Z} \sum_{d=1}^{D} \sum_{t=1}^{T} \sum_{i=1}^{T} c_{d,i} R_{d,t,i} + W$$

$$W = \sum_{t=1}^{T} \sum_{e=(u \rightarrow v)}^{E} \sum_{d'=1}^{D} \sum_{d=1}^{D} w_{e,d,d'} R_{d',t,v} Z_{d,t,u}$$

(1)

subject to Equations (2), (3), (5), (6), (7), (8), (9), (10), (11), (12), (16), (13) and (14).

*3.2.2 The Compute and Save Matrices.* $R$ is the compute matrix and $S$ the save matrix. Both matrices have the shape $|D|x|T|x|T|$.

If $R_{d,t,i} == 1$, operator $i$ is computed in timestep $t$ on device $d$. $S_{d,t,i} == 1$ means that the output of operator $i$ is saved in timestep $t$ on device $d$.

$$R_{d,t,i}, S_{d,t,i} \in \{0, 1\}$$
$$\forall d \in D, \ \forall t \in T, \ \forall i \in T$$

(2)

*3.2.3 The Free Matrix.* $F$ is the free matrix of shape $|D|x|T|x|E|$ and denotes whether tensors are freed in a timestep. If $F_{d,t,e=(u \rightarrow v)} == 1$, tensor $e = (u \rightarrow v)$ between two operators $u$ and $v$ located on device $d$ is freed in timestep $t$.

$$F_{d,t,e=(u \rightarrow v)} \in \{0, 1\}$$
$$\forall d \in D, \ \forall t \in T, \ \forall e = (u \rightarrow v) \in E$$

(3)

*3.2.4 The Availability Matrix.* $Z$ is the availability matrix of shape $|D|x|T|x|T|$ and denotes the availability of an operator output. If $Z_{d,t,i} == 1$, operator $i$ is either computed or saved in timestep $t$ on device $d$. The copy costs matrix $W$ of shape $|E|x|D|x|D|$ defines the tensor copy costs between two devices: $w_{e,d,d'}$ denotes the costs for copying the output tensor $e = (u \rightarrow v)$ of operator $u$ from device $d$ to device $d'$, since there is another operator $v$ on device $d'$ that is dependent on $u$. $Z_{d,t,u} = 1$ indicates that operator output $u$ in an edge $(u \rightarrow v)$ is computed ($R_{d,t,u} == 1$) or saved ($S_{d,t,u} == 1$) on device $d$, thus available for computing operator $v$. If the output of operator $v$ is computed, copy costs $w_{e,d,d'}$ are non-zero if the device $d$ of $u$ is not equal to the device of $v$ and thus has to be first copied before computing $v$. $Z_{d,t,u}$ ensures that copy costs are considered only once. $Z$ can be derived by the logical Equation 4:

$$Z_{d,t,u} = \begin{cases} 1 & \text{if } R_{d,t,u} == 1 \vee S_{d,t,u} == 1 \\ 0 & \text{if } R_{d,t,u} == 0 \wedge S_{d,t,u} == 0 \end{cases} \quad (4)$$

$Z_{d,t,u}$ can at maximum be 1, even if $R$ and $S$ are both 1 due to the fact that $Z$ is binary. In terms of a constraint, this can be written as Equation 5:

$$\begin{aligned} Z_{d,t,i} &\leq R_{d,t,i} + S_{d,t,i} \\ Z_{d,t,i} &\geq R_{d,t,i} \\ Z_{d,t,i} &\geq S_{d,t,i} \\ Z_{d,t,i} &\in \{0, 1\} \\ \forall d \in D, \forall t &\in T, \forall i \in T \end{aligned} \quad (5)$$

3.2.5 *The Memory Matrix.* The continuous matrix $U$ of shape $|D|x|T|x|T|$ keeps track of the memory (Equation 6). $U_{d,t,i}$ denotes the amount of occupied memory at timestep $t$ on device $d$ after computing operator $i$ and must be between 0 and $gcd_d$, where $gcd_d$ depends on the memory budget $b_d$ and ram $r_d$ of device $d$.

$$\begin{aligned} 0 \leq U_{d,t,i} &\leq gcd_d \\ gcd_d &= b_d/r_d \\ \forall d \in D, \forall t &\in T, \forall i \in T \end{aligned} \quad (6)$$

3.2.6 *The Constraints.* The following constraints are defined in the MIQP.

- *Operators are first time computed in timestep $t == i$.*
  All values in the upper right diagonal of matrix $R$ are 0 since operator $i$ is first time computed in timestep $t$ with $i == t$, which is the diagonal of the matrix. Recomputes of operator $i$ for timesteps $t' > t$ during evaluation of the new operator $i' > i$ are allowed only after the first computation of $i$ and can be located on the left lower part of the diagonal only. This can be written as Equation 7:

$$\sum_{d=1}^{D} \sum_{t=1}^{T} \sum_{i=t+1}^{T} R_{d,t,i} = 0 \quad (7)$$

- *There is one new evaluation per timestep.*
  Equation 8 ensures at least 1 new evaluation per timestep $t$ on any device $d \in D$. It defines the frontier-advancing diagonal. This constraint ensures that the new operator on the diagonal is evaluated on at least one device and, if necessary, allows for evaluation of the same operator on more than one device at the same time.

$$\sum_{d=1}^{D} R_{d,t,t} \geq 1 \ \forall t \in T \quad (8)$$

- *All operators must be evaluated at least once.*
  The sum of all new evaluations must match the total number of operators (Equation 9). Each operator must be evaluated at least once on any device in the frontier advancing diagonal.

$$\sum_{t=1}^{T} \sum_{d=1}^{D} R_{d,t,t} = T \tag{9}$$

Equation 8 and Equation 9 together state that each new evaluation should be located on exactly one device. Recomputations of tensors (lower part below the diagonal) are not restricted by this and can occur on any device.

- *No tensor can be saved before it is computed the first time.*
  Equation 10 ensures that no operator can be saved before it is computed for the first time. Thus, the upper part above the frontier advancing diagonal must be all 0.

$$\sum_{d=1}^{D} \sum_{t=1}^{T} \sum_{i=t}^{T} S_{d,t,i} = 0 \tag{10}$$

- *No tensor can be saved before it is available.*
  Equation 11 states that any operator $i$ can be saved on device $d$ in timestep $t$ only if it has been computed or saved in the previous timestep $t-1$ on the exact same device $d$. The constraint enforces that saving must take place on the same device as data are available. A tensor copy from another device to the current device followed by a save on the current device is not allowed.

$$\begin{aligned} S_{d,t+1,i} &\leq S_{d,t,i} + R_{d,t,i} \\ \forall d &\in D, \ \forall t \in T, \ \forall i \in T \end{aligned} \tag{11}$$

- *No operator can be computed before its dependencies are available.*
  Equation 12 ensures that computation of operator $v$ on device $d$ is possible only if any operator $u$ needed by $v$ is either computed or saved on any device $d' \in D$. If $d' \neq d$, copy costs will be added to the objective. In contrast to the previous constraint, this constraint allows copying a tensor ($u \rightarrow v$) from another device to the current device and to directly use it for computation of the current operator $v$.

$$\begin{aligned} R_{d,t,v} &\leq \sum_{d'=1}^{D} \left( R_{d',t,u} + S_{d',t,u} \right) \\ \forall d &\in D, \ \forall t \in T, \ \forall (u,v) \in E \end{aligned} \tag{12}$$

- *Memory initialization.*
  Equation 13 shows how the memory is initialized in the beginning of each timestep $t$. $U_{d,t,0}$ denotes the memory at the beginning of timestep $t$, which is initialized as the sum of all checkpoint memory $\sum_{i=1}^{T} S_{d,t,i} m_i$ and the memory used to hold the output of the first operator computed in the stage $m_0$.

$$\sum_{d=1}^{D} U_{d,t,0} = \sum_{i=1}^{T} S_{d,t,i} m_i + \sum_{d=1}^{D} R_{d,t,0} m_0 \quad (13)$$
$$\forall t \in T$$

- *Memory recurrence.*
  A memory recurrence constraint Equation 14 ensures that memory is allocated and freed correctly over time. After evaluating $v$, $U_{d,t,v}$ bytes of memory are in use. Before evaluating the next operator $v + 1$, the memory of $v$ and its dependencies $u \in P(v)$ can be freed if there are no further uses of $v$. If $v + 1$ is evaluated, new memory $m_{v+1}$ to store the output tensor of operator $v + 1$ has to be allocated.

$$\sum_{d=1}^{D} U_{d,t,v+1} = \sum_{d=1}^{D} U_{d,t,v} - free_{d,t,v} + R_{d,t,v+1} m_{v+1} \quad (14)$$

$free_{d,t,v}$ is the memory that is freed as soon as the output of operator $v$ is freed, including the memory of those tensor dependencies of $v$ that are only needed to compute $v$. Equation 15 reflects this: The total freed memory when $v$ is deallocated is the sum of all dependent tensors that are no longer needed after $v$ is calculated. $F_{d,t,(u \to v)} == 1$ indicates that tensor $(u \to v)$ is freed in timestep $t$. The dependencies $u \in P(v)$ are the sources of all incoming edges $e = (u \to v)$ of $v$, which means all tensors that connect a previously computed operator $u$ to $v$.

$$free_{d,t,v} = \sum_{u \in P(v) \cup \{v\}} F_{d,t,(u \to v)} \quad (15)$$
$$\forall d \in D, \ \forall t \in T$$

- *Memory limits.*
  Our $F$ matrix corresponds to the *Free* matrix of Checkmate (but with an additional dimension $D$).
  $F_{d,t,e} == 1$ denotes that tensor $e$ is freed in timestep $t$ on device $d$. Our $h(t, d, u, v)$ corresponds to *num_hazard* $(t, d, u, v)$ in Checkmate and is a linear function of decision variables. The lower and upper bounds $F_{h_{min}}$ and $F_{h_{max}}$ in Equation 16 are derived via linear reformulation of a polynomial constraint (see the Checkmate [12] paper for more details on how to derive these memory limits). We used these memory limits in our MIQP as well and adapted them to also consider the device dimension.

$$\begin{aligned}
F_{h_{min}} &\leq F_h \\
F_{h_{max}} &\geq F_h \\
-F_h + F_{h_{min}} &\leq 0 \leq -F_h + F_{h_{max}} \\
F_{h_{min}} &= \sum_{d=1}^{D} \left(1 - F_{d,t,e}\right) \\
F_{h_{max}} &= \sum_{d=1}^{D} \left(h_{max}\left(t,d,u,v\right)\left(1 - F_{d,t,e}\right)\right) \\
-F_h &= \sum_{d=1}^{D} h_{max}\left(t,d,u,v\right) \\
\forall t &\in T, \ \forall e = (u,v) \in E
\end{aligned} \quad (16)$$

### 3.3 Framework

We set up a framework using Intel's oneDNN deep learning library, which is highly optimized for Intel hardware. The network is read in as ONNX model and device budgets are defined in a first step. Tensor sizes and compute costs per network operator are obtained by computing each operator on each device. We let oneDNN choose the best memory format for each operator and are able to reorder any input tensors to the desired format of the operator, including device-to-device copies if necessary. Since oneDNN performs reordering and device copies in a single "reorder" operator, we do not model reorder costs explicitly, as they are already included in the copy costs. Next, we measure copy costs between the devices in both directions, implicitly modeling differences in reordering costs. The copy cost matrix $W$ is of $|E|$x$|D|$x$|D|$ dimensions, where $|E|$ is the number of tensor edges connecting the operator nodes.

## 4 EVALUATION

### 4.1 Setup

We use Intel's DevCloud hardware for all experiments and request nodes with the following properties:

- node "Iris"
  - Intel® Core™ i9-10920X CPU @ 3.50 GHz
    with 24 compute units, 31.05 GiB global memory
  - Intel® Iris® Xe MAX Graphics GPU @ 1.65 GHz
    with 96 compute units, 7.53 GiB global memory
- node "Gen9"
  - Intel® Xeon® E-2176G CPU @ 3.70 GHz
    with 12 compute units, 50.1 GiB global memory
  - Intel® UHD Graphics P630 GPU @ 1.2 GHz
    with 24 compute units, 62.63 GiB global memory

Since we request high-end CPUs, we simulate a lower compute performance by varying the number of threads. Although we do not consider any kind of model parallelization, the implementations of the oneDNN library are using parallelization methods internally on an operator level. Thus, we are able to achieve a lower CPU compute performance by gradually reducing the number of CPU threads. In our experiments, we vary the number of CPU threads in steps of 4 starting from 4, 8, 12, 16, 20 up to a full 24 threads. For ResNet18 and ResNet34, we also evaluate 10 threads on the "Gen9" setting.

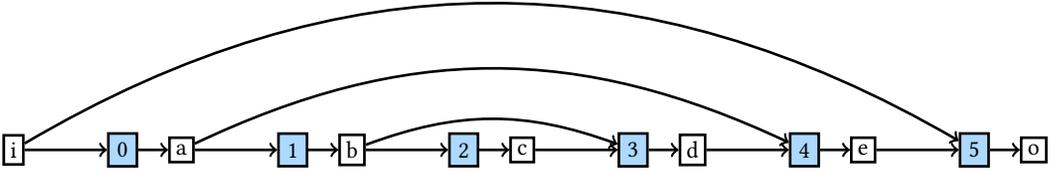

Fig. 6. Structure of our UNet. A typical block consists of: convolution, instance normalization, leaky ReLU, convolution, instance normalization, leaky ReLU. If there are 2 inputs for a block, there must be a concatenation operator right at the start of the block. Blocks 2, 3, and 4 also have one transposed convolution operator at its end. Block 5 consists only of a single convolution operator and outputs the final segmentation mask *o*.

We use the following software versions: oneDNN 2.5, Intel® oneAPI DPC++/C++ Compiler 2021.4.0 with SYCL and OpenCL 3.0, Gurobi 9.12, CBC/Cbc 2.9. We evaluate the following networks: VGG16, VGG19 [23], ResNet18, ResNet34 [9], and a UNet we defined similar to Ronneberger et al. [22] for different batchsizes. Instead of increasing the batchsize until the real device limit is reached, we consider the budget needed to save all tensors as 100%. We then artificially decrease the memory budget per device up to only 25% (75% memory reduction) of the original budget to see if our solver still finds feasible solutions and to which amount of additional compute costs.

We evaluate our XEngine solver with multiple devices against Checkmate with only one device. As for being fair, we evaluate Checkmate not only on the GPU device, as proposed in their original work, but also on the CPU device. When the budget is sufficiently large for the network and all tensors can be saved, the Checkmate schedule is exactly the same as running every operator exactly once in the corresponding device.

Our focus lies on the problem of resource selection and we do not apply any kind of model parallelization. Instead, we consider only one iteration through the network. Therefore, multiple devices do not introduce any parallelization-related benefit. In most experiments, we use one CPU and one GPU. To show that our approach can be easily extended to more than two devices, we also included experiments with one CPU and two GPUs. The two GPUs show similar compute costs due to identical hardware architecture. Since parallelization on multiple devices is not our focus in this article and the two GPUs are quite similar in runtime, we conduct most of the experiments with a device setup of one CPU and one GPU.

We evaluate various network architectures, for example, the well-known VGG and ResNet. One special network structure that is especially interesting for us is UNet, since it contains a lot of dependency structures in the form of skip connections and is at the same time compute and memory intensive due to the transposed convolution operator. Figure 6 depicts the main dependencies in the forward pass of UNet. Each "block" is a sequence of network operators consisting of two convolutions with each leaky rectified linear unit (ReLU) and instance normalization operators. The block outputs of the first half (encoder) are connected to blocks in the second half (decoder) via concatenation operators. A transposed operator is located after each decoder block.

### 4.2 Results

We evaluate our XEngine approach on various networks with different batchsizes in training and inference mode on two hardware platforms with varying numbers of CPU threads and 7 different memory budgets. For many configurations, the XEngine schedule is a single-device schedule in which all computations, including recomputations for low budgets, take place only on the fastest device. We are mostly interested in mixed schedules, in which computations take place on both the CPU and GPU. The (%) is the time reduction in % of the XEngine schedule relative to the fastest Checkmate schedule that runs only on a single device. A high negative number means a high time reduction. The best schedule varies depending on network, mode, batchsize, and number of CPU threads. When the CPU and GPU are comparably fast, the solver finds CPU/GPU schedules superior to single-device schedules. The XEngine solver runtimes vary between a few seconds for small problems (VGG16, UNet) and several hours for large problems (ResNet18). In some cases, the performance improvement is not worth running the solver: for ResNet34 with batchsize 16 in inference mode, we gain only ≤ 2% speedup on "Gen9" compared with a single-device schedule. Especially in inference mode and on linear model graphs, data dependencies of subsequent operators put higher restrictions on the schedule and allow less alternatives than strictly computing each operator one after each other. If device memory is exceeded, often the minimal cut is found where data copies are cheapest and one half is computed on the first device and the other on the second device. For other cases, such as training our UNet at a batchsize of 1, we get speedups up to 19.2% compared with running only on the GPU.

In inference mode for linear networks such as VGG, tensors are used only once to compute the next layer and always discarded without recomputation. Recomputation is especially triggered for training networks at low memory budgets and can cause the model to fit into device memory. Distributing computations also has a positive effect on inference. In all of our experiments, the solution of our XEngine solver is either the fastest single-device schedule or a faster CPU/GPU schedule. In the following, we show detailed results for the two evaluated hardware settings "Iris" and "Gen9".

*4.2.1 Iris.* We evaluate our system with different settings for inference and training of popular networks. In most settings, there is a "sweet spot" of a specific number of threads in which CPU and GPU are comparably fast and the mixed schedule is interesting: some operators are scheduled to the GPU, whereas other operators are scheduled to the CPU. In other cases—assuming sufficient memory—all operators are automatically scheduled to the fastest device. The evaluation of our approach on the "Iris" setup can be found in Table 1 with Table 1(a) for ResNet18, ResNet34 and UNet and in Table 1(b) for VGG16 and VGG19. All of the experiments in these tables are conducted with sufficient budget, focusing on the effect of operator distribution.

For VGG16 and VGG19 in inference mode, the highest speedups are achieved for running our XEngine-solver on a GPU and CPU with 12 threads. We gain up to 15.4% (VGG16) and 15.5% (VGG19) speedup compared with the fastest single-device schedule on the GPU. In training mode, 12 and 24 CPU threads yield comparable improvements up to over 10% speedup for VGG16 and VGG19.

We notice that when keeping all tensor memory, it is possible to train VGG16 and VGG19 on the GPU device up to batchsize 4 only, which needs 1.96 GiB for VGG16 and 2.08 GiB for VGG19. For ResNet18, the maximum batchsize was N=32 for the GPU, when no tensors were freed. With batchsize 128 for training ResNet18, our XEngine solver finds a CPU/GPU schedule that is 18.1% faster than the CPU Checkmate schedule by computing some operators on the GPU, whereas it cannot run on the GPU exclusively due to insufficient GPU memory. Other configurations still show the same or an improved performance compared with the fastest single-device schedule: training ResNet34 with N=128 yields a small speedup of around 5.9%. VGG16 can be trained with batchsizes N=8 to N=32 with a speedup of up to 4.5% with a mixed schedule.

Our XEngine solver finds solutions to train VGG16 on the CPU and GPU for batchsizes 8 (2.86 GiB), 16 (4.67 GiB) and even 32 (8.28 GiB), when training on the GPU alone is not possible. The

XEngine: Optimal Tensor Rematerialization 17:15

Table 1. Runtimes on "Iris" with Sufficient Budget

(a) Runtimes for ResNet18, ResNet34 and UNet on "Iris" with sufficient budget.

| | | time (ms) | | | |
|---|---|---|---|---|---|
| N | # | Checkmate CPU | Checkmate GPU | XEngine CPU+GPU (%) | budget (GiB) |
| | | ResNet18 training | | | |
| 8 | 8 | 131.9 | 204.6 | 124.9 (-5.3) | 0.61 |
| 8 | 12 | 114.0 | 203.1 | 107.3 (-5.9) | 0.61 |
| 8 | 20 | 94.0 | 207.0 | 88.25 (-6.1) | 0.61 |
| 8 | 24 | 106.1 | 209.1 | 99.9 (-5.8) | 0.61 |
| 16 | 8 | 194.7 | 310.8 | 184.5 (-5.3) | 1.13 |
| 16 | 12 | 194.3 | 312.3 | 184.2 (-5.2) | 1.13 |
| 16 | 20 | 197.1 | 313.7 | 186.3 (-5.5) | 1.13 |
| 16 | 24 | 191.7 | 314.0 | 181.8 (-5.2) | 1.13 |
| 64 | 24 | 789.9 | 1052.9 | 655.6 (-17.0) | 4.24 |
| 128 | 24 | 1595.1 | 2102.5 | 1307.0 (-18.1) | 8.38 |
| 256 | 24 | 3051.2 | 4871.4 | 2691.5 (-11.8) | 16.66 |
| | | ResNet34 training | | | |
| 8 | 8 | 184.4 | 408.92 | 173.7 (-5.8) | 0.96 |
| 8 | 12 | 209.7 | 401.7 | 198.9 (-5.2) | 0.96 |
| 8 | 20 | 175.4 | 393.9 | 162.9 (-7.1) | 0.96 |
| 8 | 24 | 182.7 | 395.2 | 172.4 (-5.7) | 0.96 |
| 16 | 20 | 352.8 | 537.7 | 333.2 (-5.5) | 1.75 |
| 128 | 24 | 2424.9 | 3112.8 | 2282.4 (-5.9) | 12.75 |
| | | UNet inference | | | |
| 1 | 12 | 28.25 | 21.7 | 19.8 (-8.9) | 0.14 |
| 1 | 24 | 24.35 | 23.7 | 19.3 (-18.5) | 0.14 |
| 2 | 12 | 52.9 | 40.8 | 37.8 (-7.4) | 0.16 |
| 2 | 24 | 43.2 | 41.4 | 34.5 (-16.8) | 0.16 |
| 4 | 12 | 86.5 | 71.1 | 63.3 (-11.0) | 0.2 |
| 4 | 24 | 72.9 | 70.5 | 57.4 (-18.5) | 0.2 |
| | | UNet training | | | |
| 1 | 12 | 61.0 | 75.36 | 49.3 (-19.2) | 0.28 |
| 1 | 24 | 55.51 | 75.3 | 45.77 (-17.6) | 0.28 |
| 2 | 12 | 109.71 | 117.5 | 90.3 (-17.7) | 0.32 |
| 2 | 24 | 103.95 | 117.19 | 85.8 (-17.5) | 0.32 |
| 4 | 12 | 211.04 | 193.8 | 169.2 (-12.7) | 0.4 |
| 4 | 24 | 187.19 | 192.96 | 155.56 (-16.9) | 0.4 |

(b) Runtimes for VGG16 and VGG19 on "Iris" with sufficient budget.

| | | time (ms) | | | |
|---|---|---|---|---|---|
| N | # | Checkmate CPU | Checkmate GPU | XEngine CPU+GPU (%) | budget (GiB) |
| | | VGG16 inference | | | |
| 2 | 12 | 54.8 | 56.6 | 46.4 (-15.4) | 0.75 |
| 2 | 24 | 44.9 | 56.3 | 44.9 (-0) | 0.75 |
| 4 | 12 | 108.8 | 94.1 | 84.6 (-10.0) | 0.97 |
| 4 | 24 | 88.8 | 94.4 | 84.3 (-5.0) | 0.97 |
| | | VGG16 training | | | |
| 2 | 12 | 195.84 | 191.38 | 171.23 (-10.5) | 1.51 |
| 2 | 24 | 194.9 | 191.77 | 171.04 (-10.8) | 1.51 |
| 4 | 12 | 353.1 | 327.3 | 308.43 (-5.8) | 1.96 |
| 4 | 24 | 343.18 | 327.3 | 307.13 (-6.2) | 1.96 |
| 8 | 12 | 678.9 | 588.4 | 565.2 (-3.9) | 2.86 |
| 8 | 24 | 641.4 | 588.6 | 563.5 (-4.3) | 2.86 |
| 16 | 12 | 1329.4 | 941.3 | 901.3 (-4.2) | 4.67 |
| 16 | 24 | 1221.5 | 941.2 | 899.3 (-4.5) | 4.67 |
| 32 | 12 | 2623.5 | 1838.8 | 1765.4 (-4.0) | 8.28 |
| 32 | 24 | 2349.4 | 1838.6 | 1759.0 (-4.3) | 8.28 |
| | | VGG19 inference | | | |
| 2 | 12 | 68.3 | 67.6 | 57.1 (-15.5) | 0.79 |
| 2 | 24 | 54.5 | 67.9 | 54.5 (-0) | 0.79 |
| 4 | 12 | 135.3 | 115.0 | 105.6 (-8.2) | 1.03 |
| 4 | 24 | 109.1 | 114.8 | 104.4 (-4.3) | 1.03 |
| | | VGG19 training | | | |
| 2 | 12 | 245.13 | 231.7 | 215.65 (-6.9) | 1.59 |
| 2 | 24 | 232.6 | 231.2 | 209.3 (-9.5) | 1.59 |
| 4 | 12 | 413.88 | 398.55 | 378.34 (-5.1) | 2.08 |
| 4 | 24 | 413.83 | 399.07 | 378.18 (-5.2) | 2.08 |
| 8 | 12 | 826.5 | 721.3 | 699.0 (-3.1) | 3.06 |
| 8 | 24 | 793.0 | 721.8 | 697.3 (-3.4) | 3.06 |
| 16 | 12 | 1631.4 | 1150.3 | 1110.3 (-3.5) | 5.0 |
| 16 | 24 | 1472.9 | 1151.0 | 1108.7 (-3.7) | 5.0 |
| 32 | 12 | 3236.6 | 2266.2 | 2193.0 (-3.2) | 8.96 |
| 32 | 24 | 2853.1 | 2254.2 | 2174.9 (-3.5) | 8.96 |

N denotes the batchsize. # denotes the number of CPU threads.

memory needed to train VGG19 is slightly higher than for VGG16. Even if the global memory on the GPU is around 7.53 GiB, the real usable free memory for our application is much smaller that that (in our experiments around 2–2.5 GiB). Figure 7c and Figure 7d visualize the results for VGG16 and VGG19 on the "Iris" setup.

The improvements for running our solver on ResNet in training mode are relatively low: the speedup we gain is around 5% to 6% compared with the fastest single-device schedule. For inference mode of ResNet, our solver schedule is as fast as the best single-device schedule. Therefore, the results are not included in Table 1(a).

Where our XEngine solver shows higher benefit is the ability to increase the overall system batchsize, higher than the maximum possible to be trained on the smaller GPU device alone and still achieve slightly better results than if we would train the network on the fastest device, if possible. When the network does not fit on the GPU device as a whole, we gain runtimes on the GPU by deallocating tensor memory after each operator evaluation regardless of dependencies, running dependent operators with dummy data.

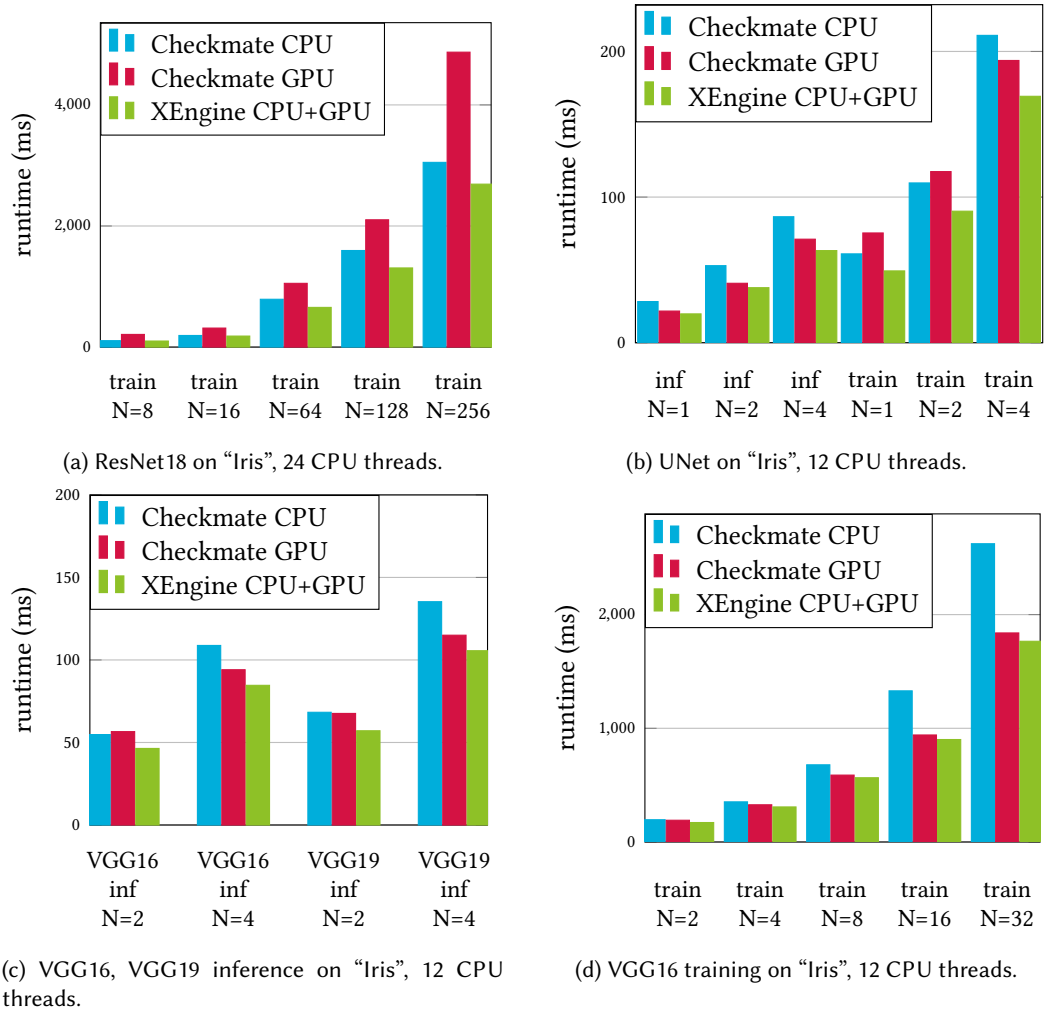

Fig. 7. Runtimes on "Iris", N=batchsize.

Visualization of the results for training ResNet18 can be found in Figure 7a. Taking a closer look at the operator runtimes, we notice that one reason for the relatively bad GPU scaling for training ResNet18 with batchsize 256 is located in the last convolution node (277 ms on the CPU and 1867 ms on the GPU) as well as the last ReLU node (64 ms on the CPU and 684 ms on the GPU). Since the last convolution is also the last operator of the network, it needs to write the operator output back to the CPU. Additionally, the original input located on the CPU is needed for backpropagation, requiring a costly reorder operation right before computation. The 10 times higher compute costs on the ReLU layer on the GPU are caused by an expensive reorder operation after the MaxPool and in the ReLU layer due to memory layout changing from 2D to 1D.

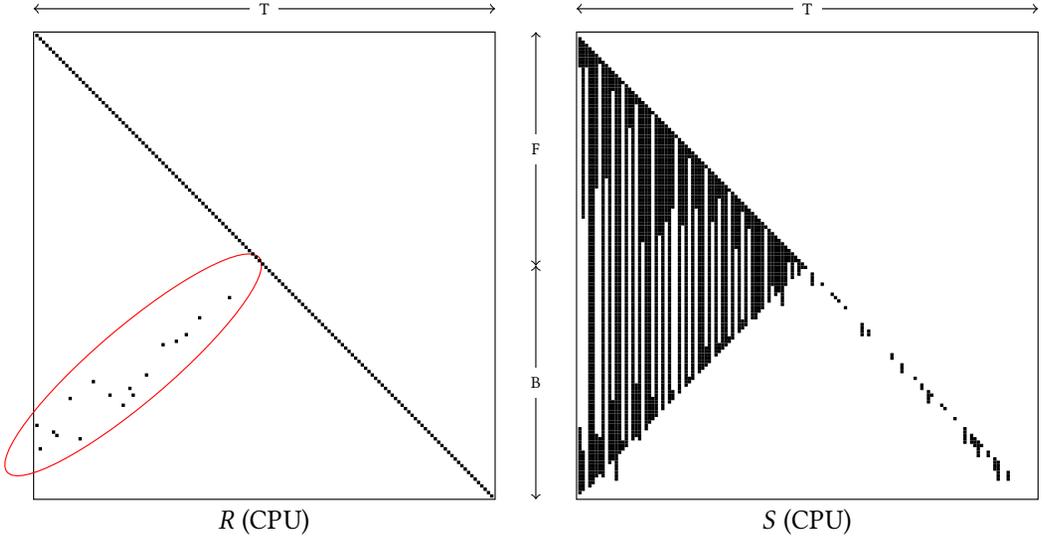

Fig. 8. Schedule for training ResNet18 with N=8 exclusively on the CPU device (24 threads) with a lower budget of 153.4 MiB (25% of full budget) on "Iris".
F=forward pass, B=backward pass, T=operators. $R$ shows 17 operator recomputations (red).

The UNet model shows great improvements in inference and training mode: our XEngine solver on a combination of CPU and GPU shows promising results of on average 13.5% less for inference and 17% less computational costs for training. The best speedup of 19.2 is achieved for training UNet with a batchsize of 1 on the GPU and a CPU with 12 threads. Figure 7b visualizes the improvement of our XEngine approach for UNet in inference and training mode when using 12 CPU threads.

In order to measure the improvement of our approach in terms of limited memory with recomputations, we vary the memory budget limit for all devices. We define a memory budget of 100% as the amount needed to save all tensors in the network, including network parameters, feature maps, and gradients.

Figure 8 shows the CPU Checkmate schedule for training ResNet18 with batchsize N=8 at a lower budget of 25%, where recomputation is triggered. The left side of the figure shows the $R$ matrix. The first line from above corresponds to the first timestep; the left upper corner indicates that operator 0 is computed in timestep 0. Starting from the top, the first half of the y axis is the forward pass; the second half corresponds to the timesteps in the backward pass. Recomputations of the forward activations are triggered in the backward step (red marked). For these points, white lines in columns of the $S$ matrix on the right side show that these operator outputs are not saved until their recomputation.

In Table 2, the ResNet and UNet architectures are trained with full and significantly lower budgets. Full budget refers to the policy that all network tensors are saved and nothing is freed. Recomputation is triggered if the budget is lower than 25% of the full budget: Additional computations occur in the backward pass as recomputation of forward activations. We compare our XEngine approach (multiple devices), denoted as CPU+GPU, against the Checkmate approach (single device) on the CPU and GPU. The additional compute costs (overhead) are denoted after every second column of lower budget compared with the full budget case. Decreasing the memory budget for training

Table 2. Runtimes for Training ResNet and UNet on "Iris" with Full and Lower Budget (25% of Full Budget)

| N | # | Checkmate CPU | time (ms) Checkmate GPU | XEngine CPU + GPU | budget (GiB) |
|---|---|---|---|---|---|
| | | | ResNet18 | | |
| 8 | 24 | 102.1 | 209.6 | 96.1 | 0.61 |
| 8 | 24 | 105.8 (+3.6%) | 217.1 (+3.6%) | 99.8 (+3.8%) | 0.15 (-75%) |
| 16 | 24 | 193.8 | 315.0 | 182.4 | 1.13 |
| 16 | 24 | 198.4 (+2.4%) | 324.9 (+3.2%) | 187.0 (+2.6%) | 0.28 (-75%) |
| | | | ResNet34 | | |
| 8 | 20 | 182.5 | 390.8 | 170.8 | 0.96 |
| 8 | 20 | 188.3 (+3.2%) | 406.4 (+4.0%) | 176.6 (+3.4%) | 0.24 (-75%) |
| 16 | 20 | 327.7 | 537.6 | 307.9 | 1.75 |
| 16 | 20 | 336.1 (+2.6%) | 557.5 (+3.7%) | 316.3 (+2.7%) | 0.44 (-75%) |
| | | | UNet | | |
| 2 | 24 | 101.9 | 116.4 | 84.3 | 0.32 |
| 2 | 24 | 110.1 (+8.1%) | 123.5 (+6.14%) | 92.5 (+9.7%) | 0.08 (-75%) |
| 4 | 24 | 189.7 | 188.8 | 158.1 | 0.4 |
| 4 | 24 | 206.0 (+8.6%) | 204.2 (+8.2%) | 174.5 (+10.3%) | 0.1 (-75%) |

N denotes the batchsize. # denotes the number of CPU threads.

ResNet34 at a batchsize of 16 to 437.2 MiB (25%) results in an additional compute cost of only 8.4 ms (+2.7%) for our XEngine schedule compared with the full budget case (307.9 ms at 1.75 GiB).

For UNet, the additional compute costs are higher than for ResNet18 and ResNet34: lowering the budget to one-quarter leads to 8.2% to 10.3% higher computational costs.

Table 3 shows the results when considering two GPU devices together with a CPU device for VGG19 and ResNet34 in training mode. With a CPU+2GPU setting on "Iris", we gain 5.8% speedup for a ResNet34 with N=16 and 20 CPU threads. Compared with 5.5% speedup using a CPU+GPU setting, this does not show any great benefit for solving our problem. The same holds for training VGG19 with a batchsize of 4 on 12 CPU threads (5.1% vs. 5.3%).

*4.2.2 Gen9.* Table 4 shows results for the "Gen9" setup. Highest speedups here can be found for VGG19 inference with up to 22.5% and VGG16 inference with up to 14.2% compared with the fastest single-device schedule. Figure 9(a) shows the results for VGG16 on the "Gen9" setup. The benefit of our solver decreases with increasing batchsize due to a bad CPU scaling. However, we can still see major performance improvements for smaller batchsizes.

For UNet, our XEngine approach leads to performance improvements of up to 17%. The results for UNet on the "Gen9" setup can be seen in Figure 9(b). This network structure shows a better scaling for the CPU than for the GPU and performance improvements are still high (16.2%) for a batchsize of 16. When trying to identify the reasons for bad scaling on the GPU for UNet on the "Gen9" setup, we noticed that the instance normalization CPU implementation was faster than the GPU version. The first and last compute blocks of UNet were also generally slower on the GPU, whereas the blocks in the middle (except instance normalization layers) did not suffer from

Table 3. Runtimes for Training of VGG19 and ResNet34 on "Iris" with Sufficient Budget on CPU and two GPUs

| N | # | Checkmate CPU | Checkmate GPU_0 | Checkmate GPU_1 | XEngine CPU + GPU_0 + GPU_1 (%) |
|---|---|---|---|---|---|
| | | | time (ms) | | |
| | | | VGG19 | | |
| 4 | 12 | 416.8 | 420.7 | 399.8 | 378.7 (-5.3) |
| | | | ResNet34 | | |
| 16 | 12 | 391.7 | 572.5 | 550.6 | 371.2 (-5.2) |
| 16 | 20 | 350.6 | 563.2 | 559.4 | 330.1 (-5.8) |
| 16 | 24 | 357.2 | 569.1 | 556.9 | 336.6 (-5.8) |

N denotes the batchsize. # denotes the number of CPU threads.

Table 4. Runtimes on "Gen9" in ms

(a) Runtimes for ResNet18, ResNet34 and UNet on "Gen9".

| N | Checkmate CPU (#) | Checkmate GPU | XEngine CPU + GPU (%) |
|---|---|---|---|
| | | ResNet18 inference | |
| 8 | 107.2 (10) | 105.6 | 100.6 (-4.7) |
| 8 | 104.2 (12) | 106.4 | 100.7 (-3.4) |
| 16 | 175.1 (10) | 186.3 | 169.6 (-3.1) |
| 32 | 364.2 (10) | 361.1 | 345.7 (-4.3) |
| 64 | 642.9 (10) | 714.2 | 632.0 (-1.7) |
| 128 | 1336.5 (10) | 1411.9 | 1313.0 (-1.8) |
| 512 | 4288.9 (10) | 5642.58 | 4288.9 (-0) |
| | | ResNet34 inference | |
| 8 | 210.4 (10) | 215.0 | 203.9 (-3.1) |
| 16 | 386.1 (10) | 359.6 | 352.7 (-1.9) |
| 32 | 727.8 (10) | 693.9 | 660.0 (-4.9) |
| 64 | 1269.2 (10) | 1373.0 | 1255.4 (-1.1) |
| 128 | 2443.17 (10) | 2864.9 | 2422.3 (-0.9) |
| 512 | 8342.75 (10) | 10825.0 | 8342.8 (-0) |
| | | UNet inference | |
| 2 | 122.27 (12) | 121.52 | 102.46 (-15.7) |
| 4 | 225.35 (10) | 229.75 | 191.95 (-14.8) |
| 4 | 195.28 (12) | 227.75 | 162.07 (-17.0) |
| 8 | 330.7 (12) | 438.0 | 275.4 (-16.7) |
| 16 | 639.9 (12) | 832.6 | 536.5 (-16.2) |

(b) Runtimes for VGG16 and VGG19 on "Gen9".

| N | Checkmate CPU (#) | Checkmate GPU | XEngine CPU + GPU (%) |
|---|---|---|---|
| | | VGG16 inference | |
| 2 | 210.5 (4) | 233.9 | 180.7 (-14.2) |
| 4 | 545.8 (4) | 411.6 | 364.9 (-11.4) |
| 8 | 1086.5 (4) | 745.5 | 702.6 (-5.8) |
| 16 | 1875.31 (4) | 1438.2 | 1398.94 (-2.7) |
| 32 | 3742.58 (4) | 2816.64 | 2786.73 (-1.0) |
| 64 | 7506.13 (4) | 5637.23 | 5565.52 (-1.3) |
| 128 | 7062.03 (10) | 11161.7 | 7062.03 (-0) |
| 256 | 14185.1 (10) | 22309.0 | 14185.1 (-0) |
| | | VGG19 inference | |
| 2 | 356.0 (4) | 320.4 | 248.2 (-22.5) |
| 4 | 694.8 (4) | 491.0 | 443.6 (-9.6) |
| 8 | 1375.7 (4) | 915.3 | 872.5 (-4.7) |
| 16 | 2372.13 (4) | 1815.23 | 1776.08 (-2.2) |
| 32 | 4736.98 (4) | 3566.1 | 3535.88 (-0.9) |
| 64 | 9472.19 (4) | 7129.88 | 7064.3 (-0.9) |
| 128 | 8882.8 (10) | 14145.2 | 8882.8 (-0) |
| 256 | 17730.4 (10) | 28341.0 | 17730.4 (-0) |

N denotes the batchsize. # denotes the number of CPU threads.

underperformance on the GPU. The first convolution needs to read the network input data from the CPU storage, which requires an expensive data copy. The last convolution operator needs to read the network input data as well to compute the backpropagation and has to write the final result

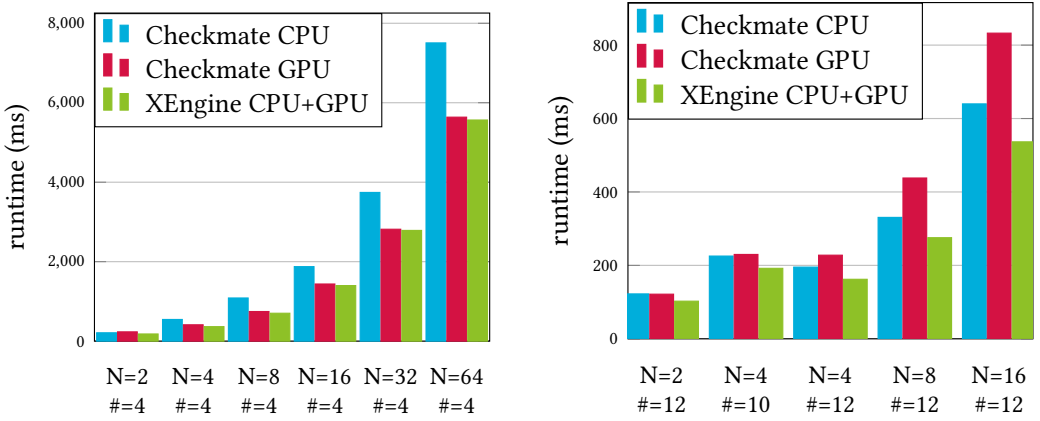

(a) Runtimes for VGG16 in inference mode on "Gen9".

(b) Runtimes for UNet in inference mode on "Gen9".

Fig. 9. Runtimes for "Gen9", N=batchsize, #=number of CPU threads.

back to storage on the CPU. The last concatenation layer of UNet and its very last convolution layer were also slower on the GPU.

## 5 DISCUSSION

### 5.1 Energy Efficiency

Our MIQP approach can be extended to other limitations depending on the hardware setup. Especially on mobile devices, energy consumption is another resource limitation that needs to be addressed. In general, we have two different ways of extending our MIQP towards this limitation. On the one hand, we can think of energy consumption as another cost term $Q$ that can be added to the original objective. We could optionally control the influence of this term by weighting it with $\alpha$. When minimizing the objective, we also find solutions that minimize the overall energy consumption but without having a guarantee of staying below a specific upper energy limit. Equation 17 shows how we could extend our MIQP with an additional weighted cost term $Q$. To obtain the energy consumption $q_{d,i}$ of operator $i$ on device $d$, we would have to acquire this information from the operating system during the offline data acquisition process.

$$\arg\min_{R,S,U,F,Z} \sum_{d=1}^{D}\sum_{t=1}^{T}\sum_{i=1}^{T} c_{d,i}R_{d,t,i} + W + \alpha Q$$

$$W = \sum_{t=1}^{T}\sum_{e=(u\to v)}^{E}\sum_{d'=1}^{D}\sum_{d=1}^{D} w_{e,d,d'}R_{d',t,v}Z_{d,t,u} \quad (17)$$

$$Q = \sum_{d=1}^{D}\sum_{t=1}^{T}\sum_{i=1}^{T} q_{d,i}R_{d,t,i}$$

subject to Equations (2), (3), (5), (6), (7), (8), (9), (10), (11), (12), (16), (13) and (14).

On the other hand, we could instead realize this by a hard constraint—in a manner similar to the upper memory limit. We can define an energy budget $Q_d^{Max}$ per device $d$ and allow those solutions that are guaranteed to be below this energy limit at any timestep $t \in T$. Equations 18 and 19 show how we could extend our MIQP with an energy constraint: Equation 18 ensures that the device-specific energy limit is not exceeded at any timestep throughout the schedule. Equation 19 ensures that the total energy costs per timestep, which also include board energy consumption, are below a maximum power limit. This term can be further refined to also model the memory subsystem in order to reflect the actual system configuration.

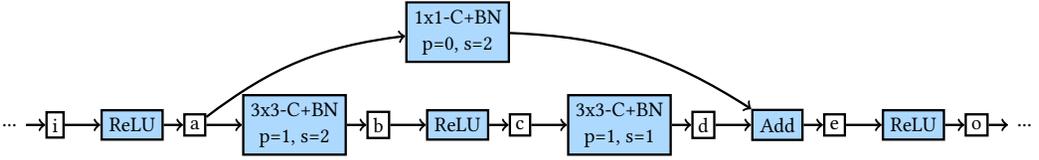

Fig. 10. Subgraph of ResNet18 showing the third skip connection connecting the 5th ReLU operator with the third add operator. C=Convolution layer with p=padding and s=stride. BN=Batch normalization layer.

$$0 \leq Q_{d,t,i} \leq Q_d^{Max}$$
$$\forall d \in D, \ \forall t \in T, \ \forall i \in T \quad (18)$$

$$\sum_{d=1}^{D} \sum_{i=1}^{T} Q_{d,t,i} + Q_{Board} \leq Q^{Max}$$
$$\forall d \in D, \ \forall t \in T, \ \forall i \in T \quad (19)$$

Since we did not find a way to easily access power consumption information on the Intel DevCloud, our result section is lacking experiments for an extended MIQP that also optimizes energy efficiency.

### 5.2 Model Parallelization

Model parallelization is often understood as dividing a network graph into groups of layers and computing them separately on different devices in parallel. Our MIQP outputs a schedule that decides on which device a layer should be computed and also considers copy costs for data transfers. Nevertheless, we do not consider parallel execution of layers. Rather, we assume that the operators are executed one after the other. The benefit introduced by model parallelization heavily depends on network architecture. Linear graphs do not benefit much from model parallelization, since subsequent layers depend on each other. Instead, it is better to divide the data into batches and to make use of data parallelization. If skip connections are present—as for UNet or ResNet—one may introduce model parallelization and compute concurrent branches in parallel. Skip connections often contain two "branches": the inner, which is the more compute-intensive branch, and the outer branch, often only a data dependency between two operators. In Figure 6, three outer branches are depicted that themselves do not contain any computation at all. Figure 10 shows a ResNet18 skip connection with two branches. In the inner branch, two 3 × 3 convolutions, each followed by a batch normalization layer, are connected by a ReLU layer. The inner branch contains even more than twice the computational demand compared with the outer branch, which contains only one 1 × 1 convolution layer followed by a batch normalization layer. The wall-clock time depends on the branch, which, on average, takes longer to compute. Since the outer skip branches of skip connections are often computationally cheap, the bottleneck is significantly more often the memory transfer rather than the computation of the outer skip branch. Executing the branches in parallel will not achieve notable performance speedup.

### 5.3 Alternative Platforms

Our experiments were conducted on Intel hardware and software since we use Intel oneDNN as a backend. Nevertheless, the key concepts of our approach can be transferred to other platforms, such as AMD and NVIDIA. Intel is working on an NVIDIA backend for oneDNN that is still in experimental mode at this writing. According to the oneDNN GitHub repository, Intel will restructure the engine creation with the goal of integrating the NVIDIA backend more tightly into the runtime.

## 6 CONCLUSION

This article presents XEngine, an MIQP that extends the Checkmate MILP to multiple heterogeneous devices, combining distributed computation with checkpointing. We use additional constraints to hand computation over between different devices with the accompanying memory copy costs. We find optimal schedules according to a given memory budget per device and recompute operator outputs whenever the model does not meet memory limitations. We evaluated our approach against Checkmate on several networks—such as ResNet, VGG, and UNet—for inference and training. In contrast to Checkmate, we also distribute the network operators between a CPU and a GPU. This way, we gain up to 19.2% speedup for training a UNet with a CPU/GPU schedule and a speedup of up to 22.5% for inference of a VGG19 compared with computing only on the GPU. Moreover, we can overcome the problem that some networks cannot be exclusively computed on the GPU by CPU/GPU schedules, such as for ResNet18 with batchsize 128 on the Iris Xe MAX GPU. Despite all of our experiments being conducted on Intel hardware, our approach can be transferred to other platforms as well, such as AMD or NVIDIA hardware.

## ACKNOWLEDGMENTS

We would like to thank Intel for providing access to their DevCloud environment for evaluation. Additionally, we would like to thank Olivier Beaumont and Denise Cucchiara for constructive feedback on our paper.

## REFERENCES

[1] Olivier Beaumont, Lionel Eyraud-Dubois, Julien Herrmann, Alexis Joly, and Alena Shilova. 2019. Optimal Checkpointing for Heterogeneous Chains: How to Train Deep Neural Networks with Limited Memory. *CoRR* abs/1911.13214 (2019), 1–27. arXiv:1911.13214

[2] Olivier Beaumont, Lionel Eyraud-Dubois, and Alena Shilova. 2021. Efficient Combination of Rematerialization and Offloading for Training DNNs. In *Advances in Neural Information Processing Systems (NeurIPS)*, Marc'Aurelio Ranzato, Alina Beygelzimer, Yann N. Dauphin, Percy Liang, and Jennifer Wortman Vaughan (Eds.). Curran Associates, Inc., 23844–23857.

[3] Tianqi Chen, Bing Xu, Chiyuan Zhang, and Carlos Guestrin. 2016. Training Deep Nets with Sublinear Memory Cost. *CoRR* abs/1604.06174 (2016), 1–12. arXiv:1604.06174

[4] Aidan N. Gomez, Mengye Ren, Raquel Urtasun, and Roger B. Grosse. 2017. The Reversible Residual Network: Backpropagation without Storing Activations. In *Proceedings of the 31st International Conference on Neural Information Processing Systems (NIPS)*, Isabelle Guyon, Ulrike von Luxburg, Samy Bengio, Hanna M. Wallach, Rob Fergus, S. V. N. Vishwanathan, and Roman Garnett (Eds.). Curran Associates Inc., 2211–2221.

[5] Andreas Griewank. 1992. Achieving Logarithmic Growth of Temporal and Spatial Complexity in Reverse Automatic Differentiation. *Optimization Methods and Software* 1, 1 (1992), 35–54. https://doi.org/10.1080/10556789208805505

[6] Andreas Griewank and Andrea Walther. 2000. Algorithm 799: Revolve: An Implementation of Checkpointing for the Reverse or Adjoint Mode of Computational Differentiation. *ACM Trans. Math. Software* 26, 1 (mar 2000), 19–45. https://doi.org/10.1145/347837.347846

[7] Audrūnas Gruslys, Rémi Munos, Ivo Danihelka, Marc Lanctot, and Alex Graves. 2016. Memory-Efficient Backpropagation through Time. In *Proceedings of the 30th International Conference on Neural Information Processing Systems (NIPS)*, Daniel D. Lee, Masashi Sugiyama, Ulrike von Luxburg, Isabelle Guyon, and Roman Garnett (Eds.). Curran Associates Inc., 4132–4140.

[8] Laurent Hascoët and Valérie Pascual. 2013. The Tapenade Automatic Differentiation Tool: Principles, Model, and Specification. *ACM Trans. Math. Software* 39, 3 (2013), 20:1–20:43. https://doi.org/10.1145/2450153.2450158

[9] Kaiming He, Xiangyu Zhang, Shaoqing Ren, and Jian Sun. 2016. Deep Residual Learning for Image Recognition. In *2016 IEEE Conference on Computer Vision and Pattern Recognition (CVPR)*. IEEE, 770–778. https://doi.org/10.1109/CVPR.

2016.90

[10] Zhongzhe Hu, Junmin Xiao, Zheye Deng, Mingyi Li, Kewei Zhang, Xiaoyang Zhang, Ke Meng, Ninghui Sun, and Guangming Tan. 2022. MegTaiChi: Dynamic Tensor-Based Memory Management Optimization for DNN Training. In *Proceedings of the 36th ACM International Conference on Supercomputing (ICS)*, Lawrence Rauchwerger, Kirk W. Cameron, Dimitrios S. Nikolopoulos, and Dionisios N. Pnevmatikatos (Eds.). ACM, 25:1–25:13. https://doi.org/10.1145/3524059.3532394

[11] Chien-Chin Huang, Gu Jin, and Jinyang Li. 2020. SwapAdvisor: Pushing Deep Learning Beyond the GPU Memory Limit via Smart Swapping. In *Proceedings of the 25th International Conference on Architectural Support for Programming Languages and Operating Systems (ASPLOS)*, James R. Larus, Luis Ceze, and Karin Strauss (Eds.). ACM, 1341–1355. https://doi.org/10.1145/3373376.3378530

[12] Paras Jain, Ajay Jain, Aniruddha Nrusimha, Amir Gholami, Pieter Abbeel, Joseph Gonzalez, Kurt Keutzer, and Ion Stoica. 2020. Checkmate: Breaking the Memory Wall with Optimal Tensor Rematerialization. In *Proceedings of Machine Learning and Systems (MLSys)*, Inderjit S. Dhillon, Dimitris S. Papailiopoulos, and Vivienne Sze (Eds.). mlsys.org, 497–511.

[13] Haotian Jiang, James Starkman, Yu-Ju Lee, Huan Chen, Xiaoye Qian, and Ming-Chun Huang. 2021. Distributed Deep Learning Optimized System over the Cloud and Smart Phone Devices. *IEEE Transactions on Mobile Computing (TMC)* 20, 1 (2021), 147–161. https://doi.org/10.1109/TMC.2019.2941492

[14] Marisa Kirisame, Steven Lyubomirsky, Altan Haan, Jennifer Brennan, Mike He, Jared Roesch, Tianqi Chen, and Zachary Tatlock. 2021. Dynamic Tensor Rematerialization. In *9th International Conference on Learning Representations (ICLR)*. OpenReview.net, 1–31.

[15] Ravi Kumar, Manish Purohit, Zoya Svitkina, Erik Vee, and Joshua Wang. 2019. Efficient Rematerialization for Deep Networks. In *Advances in Neural Information Processing Systems (NeurIPS)*, Hanna M. Wallach, Hugo Larochelle, Alina Beygelzimer, Florence d'Alché-Buc, Emily B. Fox, and Roman Garnett (Eds.). Curran Associates, Inc., 15146–15155.

[16] Mitsuru Kusumoto, Takuya Inoue, Gentaro Watanabe, Takuya Akiba, and Masanori Koyama. 2019. A Graph Theoretic Framework of Recomputation Algorithms for Memory-Efficient Backpropagation. In *Advances in Neural Information Processing Systems (NeurIPS)*, Hanna M. Wallach, Hugo Larochelle, Alina Beygelzimer, Florence d'Alché-Buc, Emily B. Fox, and Roman Garnett (Eds.). Curran Associates, Inc., 1161–1170.

[17] Jianjin Liao, Mingzhen Li, Qingxiao Sun, Jiwei Hao, Fengwei Yu, Shengdong Chen, Ye Tao, Zicheng Zhang, Hailong Yang, Zhongzhi Luan, and Depei Qian. 2022. Mimose: An Input-Aware Checkpointing Planner for Efficient Training on GPU. *CoRR* abs/2209.02478 (2022), 1–13. arXiv:2209.02478

[18] Yanchen Liu, Myung J. Lee, and Yanyan Zheng. 2016. Adaptive Multi-Resource Allocation for Cloudlet-Based Mobile Cloud Computing System. *IEEE Transactions on Mobile Computing (TMC)* 15, 10 (2016), 2398–2410. https://doi.org/10.1109/TMC.2015.2504091

[19] Shishir G. Patil, Paras Jain, Prabal Dutta, Ion Stoica, and Joseph E. Gonzalez. 2022. POET: Training Neural Networks on Tiny Devices with Integrated Rematerialization and Paging. In *Proceedings of the 39th International Conference on Machine Learning (ICML)*, Kamalika Chaudhuri, Stefanie Jegelka, Le Song, Csaba Szepesvári, Gang Niu, and Sivan Sabato (Eds.). PMLR, 17573–17583.

[20] Xuan Peng, Xuanhua Shi, Hulin Dai, Hai Jin, Weiliang Ma, Qian Xiong, Fan Yang, and Xuehai Qian. 2020. Capuchin: Tensor-Based GPU Memory Management for Deep Learning. In *Proceedings of the 25th International Conference on Architectural Support for Programming Languages and Operating Systems (ASPLOS)*, James R. Larus, Luis Ceze, and Karin Strauss (Eds.). ACM, 891–905. https://doi.org/10.1145/3373376.3378505

[21] Jie Ren, Samyam Rajbhandari, Reza Yazdani Aminabadi, Olatunji Ruwase, Shuangyan Yang, Minjia Zhang, Dong Li, and Yuxiong He. 2021. ZeRO-Offload: Democratizing Billion-Scale Model Training. *CoRR* abs/2101.06840 (2021), 1–14. arXiv:2101.06840

[22] Olaf Ronneberger, Philipp Fischer, and Thomas Brox. 2015. U-Net: Convolutional Networks for Biomedical Image Segmentation. In *Medical Image Computing and Computer-Assisted Intervention (MICCAI)*, Nassir Navab, Joachim Hornegger, William M. Wells III, and Alejandro F. Frangi (Eds.). Springer, 234–241. https://doi.org/10.1007/978-3-319-24574-4_28

[23] Karen Simonyan and Andrew Zisserman. 2015. Very Deep Convolutional Networks for Large-Scale Image Recognition. In *3rd International Conference on Learning Representations (ICLR)*, Yoshua Bengio and Yann LeCun (Eds.). 1–14.

[24] Jeffrey Siskind and Barak Pearlmutter. 2018. Divide-and-Conquer Checkpointing for Arbitrary Programs with No User Annotation. *Optimization Methods and Software* 33, 4-6 (2018), 1288–1330. https://doi.org/10.1080/10556788.2018.1459621

[25] Yu Tang, Chenyu Wang, Yufan Zhang, Yuliang Liu, Xingcheng Zhang, Linbo Qiao, Zhiquan Lai, and Dongsheng Li. 2022. DELTA: Dynamically Optimizing GPU Memory beyond Tensor Recomputation. *CoRR* abs/2203.15980 (2022), 1–12. arXiv:2203.15980

[26] Duc Van Le and Chen-Khong Tham. 2018. A Deep Reinforcement Learning based Offloading Scheme in ad-hoc Mobile Clouds. In *IEEE Conference on Computer Communications Workshops (INFOCOM WKSHPS)*. IEEE, 760–765. https://doi.org/10.1109/INFCOMW.2018.8406881

[27] Linnan Wang, Jinmian Ye, Yiyang Zhao, Wei Wu, Ang Li, Shuaiwen Leon Song, Zenglin Xu, and Tim Kraska. 2018. Superneurons: Dynamic GPU Memory Management for Training Deep Neural Networks. In *Proceedings of the 23rd ACM SIGPLAN Symposium on Principles and Practice of Parallel Programming (PPoPP)*, Andreas Krall and Thomas R. Gross (Eds.). ACM, 41–53. https://doi.org/10.1145/3200691.3178491

[28] Lijie Wen, Zan Zong, Li Lin, and Leilei Lin. 2022. A Swap Dominated Tensor Re-Generation Strategy for Training Deep Learning Models. In *2022 IEEE International Parallel and Distributed Processing Symposium (IPDPS)*. IEEE, 996–1006. https://doi.org/10.1109/IPDPS53621.2022.00101

## A APPENDIX

Figures 11 to 18 depict selected schedules of the "Iris" setup in Table 2. Figures 11 and 12 show the schedules for ResNet18 of our XEngine approach on a CPU-GPU setup (Figure 11) in comparison with CPU-Checkmate (Figure 12, left two images) and GPU-Checkmate (Figure 12, right two images).

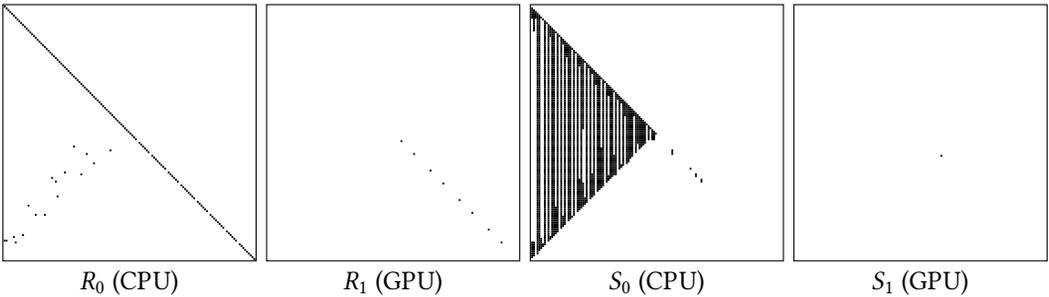

Fig. 11. XEngine schedule for training ResNet18 on "Iris" with N=8. Budget: 153.4 MiB (25%), 24 CPU threads: $R_0$, $R_1$, $S_0$, $S_1$, Table 2 line 2.

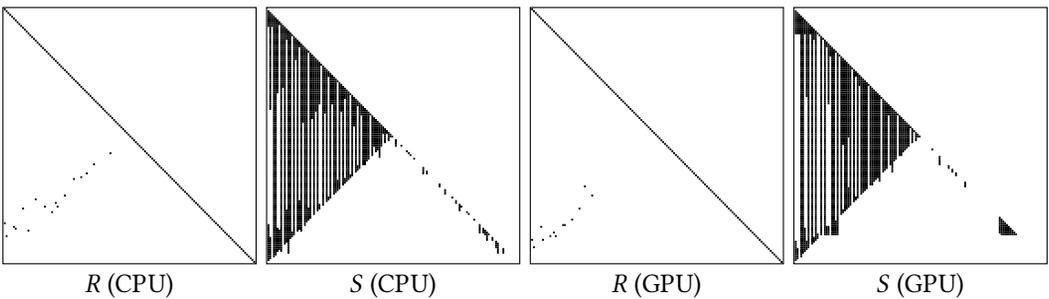

Fig. 12. Checkmate schedules for training ResNet18 on "Iris" with N=8. Budget: 153.4 MiB (25%), 24 CPU threads, CPU: $R$, $S$, GPU: $R$, $S$, Table 2 line 2.

Other figures show the schedules for ResNet34 (Figures 13 and 14) and UNet (Figures 15 to 18).
For UNet, we selected two schedule comparisons, one for batchsize 2 (Figures 15 and 16) and one for batchsize 4 (Figures 17 and 18).

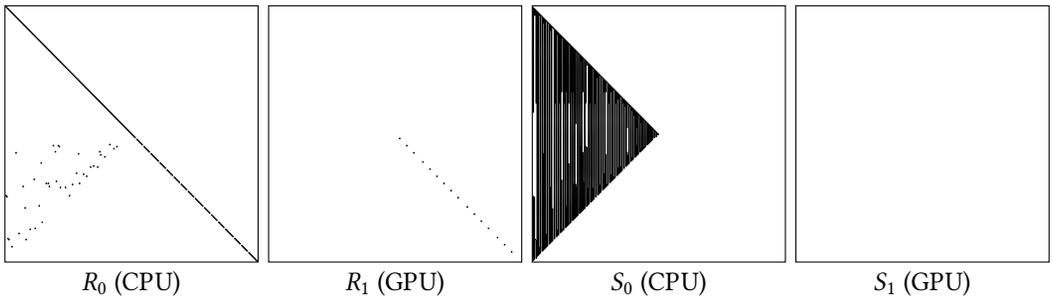

Fig. 13. XEngine schedule for training ResNet34 on "Iris" with N=8. Budget: 240.8 MiB (25%), 20 CPU threads: $R_0$, $R_1$, $S_0$, $S_1$, Table 2 line 6.

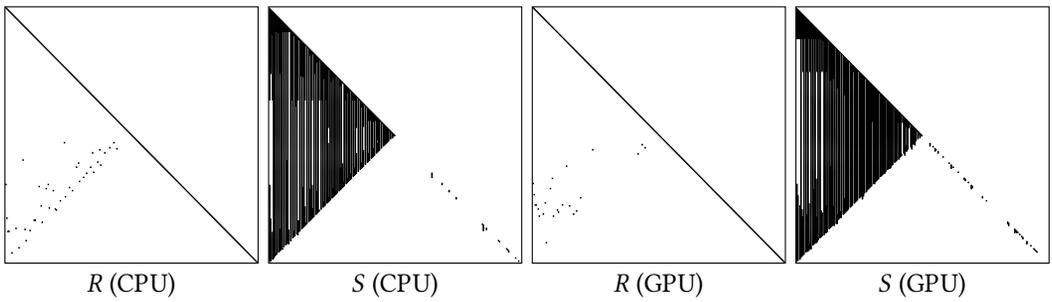

Fig. 14. Checkmate schedules for training ResNet34 on "Iris" with N=8. Budget: 240.8 MiB (25%), 20 CPU threads, CPU: $R$, $S$, GPU: $R$, $S$, Table 2 line 6.

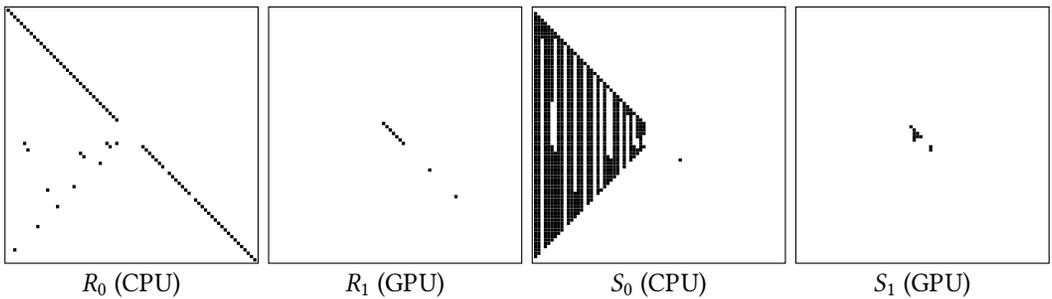

Fig. 15. XEngine schedule for training UNet on "Iris" with N=2. Budget: 80 MiB (25%), 24 CPU threads: $R_0$, $R_1$, $S_0$, $S_1$, Table 2 line 10.

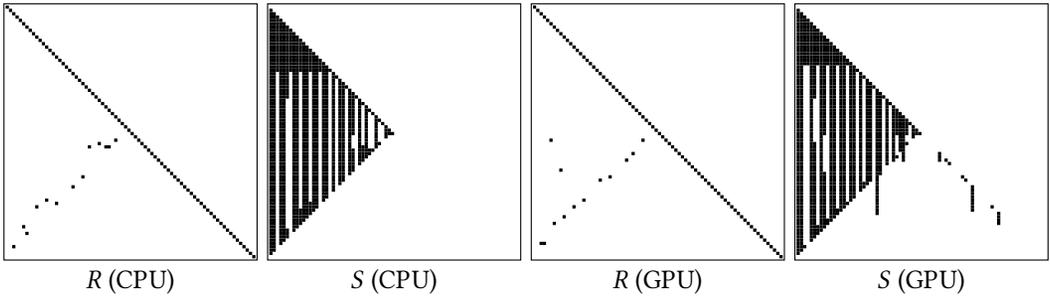

Fig. 16. Checkmate schedules for training UNet on "Iris" with N=2. Budget: 80 MiB (25%), 24 CPU threads, CPU: $R$, $S$, GPU: $R$, $S$, Table 2 line 10.

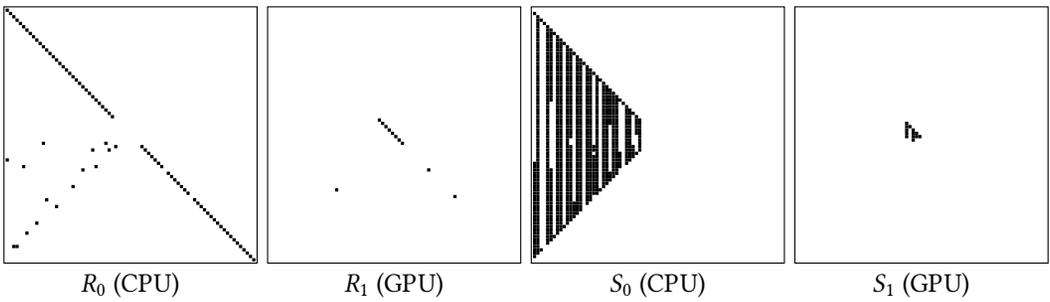

Fig. 17. XEngine schedule for training UNet on "Iris" with N=4. Budget: 99.3 MiB (25%), 24 CPU threads: $R_0$, $R_1$, $S_0$, $S_1$, Table 2 line 12.

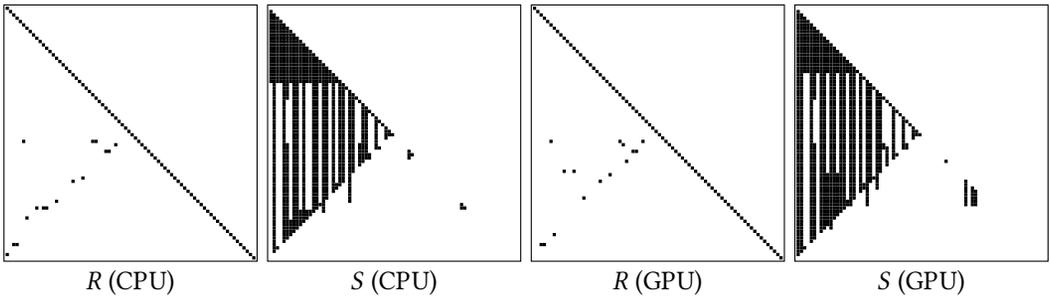

Fig. 18. Checkmate schedules for training UNet on "Iris" with N=4. Budget: 99.3 MiB (25%), 24 CPU threads, CPU: $R$, $S$, GPU: $R$, $S$, Table 2 line 12.